\documentclass{article}

\ifdefined\XeTeXversion
  \usepackage[T1]{fontenc}
\fi
\usepackage{microtype}
\usepackage{graphicx}
\usepackage{subcaption}
\usepackage{booktabs} 



\usepackage{hyperref}
\ifdefined\XeTeXversion
  \usepackage{xurl}
\fi

\usepackage[preprint]{icml2027}

\usepackage{amsmath}
\usepackage{amssymb}
\usepackage{mathtools}
\usepackage{amsthm}

\usepackage[capitalize,noabbrev]{cleveref}

\theoremstyle{plain}

\theoremstyle{definition}

\theoremstyle{remark}

\usepackage[textsize=tiny]{todonotes}

\icmltitlerunning{SWGDN}

\begin{document}

\twocolumn[
  \icmltitle{SWGDN: Improving Gated Delta Networks with Sliding-Window Attention}



  \icmlsetsymbol{equal}{*}

  \begin{icmlauthorlist}
    \icmlauthor{Yuchen Song}{}
    
\end{icmlauthorlist}

\icmlcorrespondingauthor{ysong8162@gmail.com}

  \icmlkeywords{Linear Attention,DeltaNet}

  \vskip 0.3in
]

\makeatletter
\global\icml@noticeprintedtrue
{\let\thefootnote\relax
 \footnotetext{\Notice@String}}
\makeatother





\begin{abstract}

Linear attention compresses historical context into a fixed-size recurrent state, reducing the computational and memory costs of long-sequence modeling. From the perspective of test-time regression (TTR), the state matrix can be viewed as an online-learned regression model, where keys and values serve as inputs and targets respectively, and queries are used to read from the model. From this perspective, the effectiveness of state learning depends not only on the update rule but also on how the key–value learning pairs used for state updates are constructed. While existing work has largely focused on how the state is updated and organized, we investigate how the learning pairs used for state updates are constructed. To this end, we propose \textbf{S}liding-\textbf{W}indow \textbf{G}ated \textbf{D}elta\textbf{N}et (SWGDN), which replaces short convolutions in the Delta-Rule linear attention with sliding-window attention to perform content-dependent local aggregation of projected queries, keys, and values. Our experiments show that SWGDN makes the key--value pairs used in consecutive updates more similar than the baseline. The local mixing strategy can be incorporated into any Delta-Rule-based linear attention architectures without changing their state update equations or chunkwise parallel algorithms. We apply this strategy to two Delta-Rule variants, GDN2 and KDA at the 340M-parameter scale, experimental results demonstrate that SWGDN outperforms the original Delta-Rule baseline across multiple benchmarks.

\end{abstract}

\section{Introduction}

Transformer \cite{NIPS2017_3f5ee243} have become the dominant architecture for language modeling due to their explicit access to historical context and efficient parallel computation. However, the computational complexity of standard full attention grows quadratically with sequence length, while autoregressive decoding requires an increasingly large KV cache, introducing substantial computational and memory overhead for long-context modeling. In contrast, linear attention~\cite{
  katharopoulos_transformers_2020,
  choromanski2021performer,
  peng2021random,
  kasai2021finetuning,
  peng2023rwkvreinventingrnnstransformer,
  sun2023retentivenetworksuccessortransformer,
  yang2024gatedlinearattentiontransformers,
  qin2024hgrn2} compresses historical information into a fixed-size recurrent state, reducing sequence-processing complexity to linear time and enabling constant per-token computational and memory costs during decoding with respect to context length.

Test-Time Regression (TTR) ~\cite{wang_test-time_2026} provides a perspective on linear attention, interpreting the state matrix $S$ as a regression model that learns the mapping $Sk=v$. At each position, the key-value pair $(k_t,v_t)$ forms a training example, while the query $q_t$ retrieves information from the current state through $S_tq_t$. From this perspective, different linear attention architectures can be understood as employing different optimization strategies. Linear attention corresponds to a batch gradient step from zero, while DeltaNet~\cite{yang_parallelizing_2025,schlag_linear_2021} performs a single online stochastic gradient descent step to correct the current prediction. GDN and KDA~\cite{kimiteam2025kimilinearexpressiveefficient,yang2025gateddeltanetworksimproving,gu2024mambalineartimesequencemodeling} further introduce forgetting mechanisms and channel-wise decay, whereas GDN2 ~\cite{hatamizadeh_gated_2026} decouples erase and write gates, which admits a preconditioned-gradient interpretation. Following this direction, MesaNet~\cite{oswald_mesanet_2026,vonoswald2024uncoveringmesaoptimizationalgorithmstransformers} and Preconditioned Gated DeltaNet (PGDN)~ \cite{tumma_preconditioned_2026} improve regression fitting through conjugate gradient methods and curvature-aware preconditioning, respectively. However, these approaches primarily focus on how to solve the regression problem, explicitly modifying the state update equations and introducing more complex chunkwise parallel algorithms. This raises a complementary question: \textbf{Can we improve linear attention while preserving the original recurrent update rules and chunkwise parallel algorithms?}

One promising direction is to improve the key–value pairs involved in each recurrent update. In natural language sequences, many tokens correspond to relatively uninformative key–value pairs, and writing these pairs into the state matrix can interfere with previously stored associations. Although a series of prior works have introduced mechanisms such as forget gates~\cite{dao2024transformersssmsgeneralizedmodels,sun2023retentivenetworksuccessortransformer,peng2024eaglefinchrwkvmatrixvalued} to remove irrelevant information from the state, we argue that these approaches do not address the problem at its source. A more fundamental approach is to endow the model with the ability to selectively determine what information to write into the state. Specifically, less informative tokens should receive lower weights during local mixing, thereby reducing their interference with existing associations, while more informative tokens should receive higher weights.This intuition naturally aligns with the attention mechanism: tokens within a local window are scored through content-dependent dot products and compete for attention weights through softmax. When the model considers a token less informative, it can assign a smaller weight to that token. Conversely, more informative tokens can receive larger weights and thus contribute more substantially to the locally aggregated representation.

Existing local mixing mechanisms typically employ channel-wise short convolutions. However, a fundamental limitation of this approach is that the aggregation weights are fixed after training and cannot adapt to the content within each local window. Although recent studies have explored replacing static convolutions with dynamic convolutions\cite{sieberling2026dynamicshortconvolutionsimprove,wu2019payattentionlightweightdynamic}, such approaches do not necessarily enable explicit content-based selection among tokens within the window and lack the explicit competitive mechanism inherent in attention.

We propose \textbf{S}liding-\textbf{W}indow \textbf{G}ated \textbf{D}eltaNet (SWGDN), which replaces short convolutions with sliding-window attention. To reduce parameter count and computational overhead, each branch (Q, K, and V) uses learned queries shared across positions to score the projected keys within its window. Attention enables the model to adaptively assign weights based on the content within each window, downweighting irrelevant key–value pairs while upweighting important ones.

Importantly, our modification operates only on local mixing before the recurrent update, allowing the original recurrence rules and chunkwise parallel algorithms to be reused without modification. We compare three attention weight-sharing schemes: sharing weights across all channels within each head, sharing weights within groups of four channels, and assigning independent weights to each channel. Our contributions are summarized as follows.
\begin{itemize}
  \item \textbf{We establish learning-pair construction as a design axis for linear attention.} SWGDN uses sliding-window attention to construct the keys and values that serve as regression inputs and targets. It also mixes queries for state readout and reuses the original recurrent update rule and chunk-parallel kernels.
  \item \textbf{We show that changing local key and value mixing can improve retrieval.} On GDN2, applying sliding-window attention only to K and V raises mean RULER NIAH accuracy from 35.0\% to 44.9\%, with the original Q-branch convolution and recurrent update rule in place (\cref{tab:niah,tab:local_mixing_ablation}).

\end{itemize}

\begin{figure*}[tbp]
\centering
\includegraphics[width=\textwidth]{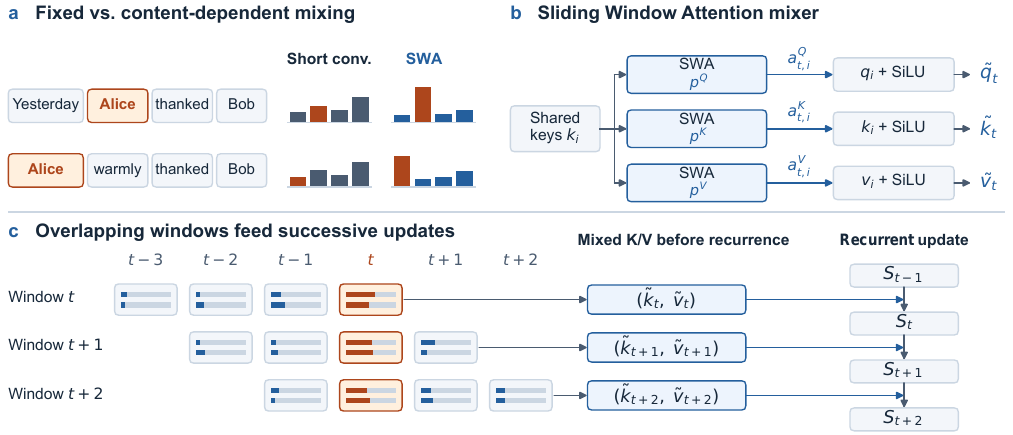}
\caption{\textbf{Sliding-Window Gated DeltaNet (SWGDN).}
\textbf{(a)} Short convolutions use fixed coefficients at each relative position, while sliding-window attention can emphasize a token at different offsets.
\textbf{(b)} In the G1 configuration shown, each branch uses one learned query per head (\(p^Q\), \(p^K\), or \(p^V\)).
All three branches score the same projected keys and use their own softmax weights to mix Q, K, or V.
Here, \(i\) indexes an input token within the sliding window, and \(t\) denotes the position at which the mixed output is produced.
\textbf{(c)} A token shared across overlapping windows can contribute to several mixed K/V pairs that feed the original gated updates.}
\label{fig:swgdn_method}
\end{figure*}

\section{Background}

\subsection{Test-Time Regression}

Test-Time Regression (TTR) interprets the state matrix $S$ in linear attention as an online regression model. Specifically, key-value pairs $(k_t, v_t)$ serve as input samples and regression targets, respectively. The model stores their associations by learning the mapping $Sk_t \approx v_t$ and retrieves stored information through $o_t = S_t q_t$. From this perspective, the state accumulation in linear attention can be interpreted as a single batch gradient descent step on the squared error over historical samples, starting from zero initialization. For the current training pair $(k_t, v_t)$, the least-squares objective is defined as
\begin{equation}
\ell_t(S) = \frac{1}{2}\|Sk_t - v_t\|_2^2.
\end{equation}
In contrast, DeltaNet performs a single online gradient descent step on the current sample, starting from the previous state $S_{t-1}$:

\begin{equation}
\begin{aligned}
S_t &= S_{t-1} - \beta_t \nabla_S \ell_t(S_{t-1}) \\
    &= S_{t-1} + \beta_t (v_t - S_{t-1}k_t)k_t^\top.
\end{aligned}
\end{equation}

Gated DeltaNet (GDN) introduces data-dependent decay, while Kimi Delta Attention (KDA) extends decay to individual channels. With $D_t=\operatorname{Diag}(\alpha_t)$ and update rate $\beta_t$, KDA updates its state as
\begin{equation}
\begin{aligned}
\bar{S}_t &= S_{t-1}D_t, \\
S_t &= \bar{S}_t + \beta_t(v_t-\bar{S}_t k_t)k_t^\top.
\end{aligned}
\end{equation}
Gated DeltaNet-2 (GDN2) retains channel-wise decay and decouples the key-side erasure gate $b_t$ from the value-side write gate $w_t$ :
\begin{equation}
\begin{aligned}
\bar{S}_t &= S_{t-1}D_t, \\
S_t &= \bar{S}_t + (z_t - \bar{S}_t e_t)k_t^\top, \\
e_t &= b_t \odot k_t, \qquad
z_t = w_t \odot v_t.
\end{aligned}
\end{equation}
This update admits a preconditioned gradient interpretation on the gated learning pair $(e_t,z_t)$. \Cref{app:linear_attention_derivation} provides further background on recurrent linear attention and gated delta updates.

\subsection{Sliding Window Attention}

Sliding Window Attention (SWA)\cite{beltagy_longformer_2020} is a local attention mechanism designed to reduce the computational and memory overhead of standard self-attention in long-sequence modeling. Unlike standard causal self-attention, which allows each query to attend to all preceding tokens, SWA restricts attention to a fixed-size local window. Specifically, for the query vector $q_t$ at position $t$, the attention output is defined as
\begin{equation}
o_t = \sum_{j=\max(1,t-W+1)}^{t} \alpha_{t,j}v_j,
\end{equation}

where the attention weights are computed as
\begin{equation}
\alpha_{t,j} =
\frac{\exp(q_t^\top k_j / \sqrt{d_k})}
{\sum_{i=\max(1,t-W+1)}^{t}
\exp(q_t^\top k_i / \sqrt{d_k})}.
\end{equation}
Here, $W$ denotes the window size, $k_j$ and $v_j$ represent the key and value vectors at position $j$, respectively, and $d_k$ is the key dimension. By restricting the attention range of each query, SWA reduces the attention computation complexity from $O(N^2)$ to $O(NW)$ for a sequence of length $N$. Unlike short causal convolutions, which aggregate local information using fixed convolutional kernels, SWA dynamically assigns attention weights based on query-key similarity, enabling content-dependent interactions within the local window\cite{han2021localattention}.

\section{Sliding Window Gated DeltaNet}

In Gated DeltaNet, short convolutions mix local context into the projected queries, keys, and values before recurrent updates and memory retrieval. From the TTR perspective, the resulting keys and values serve as regression inputs and targets for state updates, making local mixing part of the construction of recurrent training pairs. However, short convolutions assign mixing coefficients according to relative position, independently of the current window content. For example, consider “Yesterday [Alice] [thanked] [Bob]” and “[Alice] warmly [thanked] [Bob].” When constructing the representation at “Bob,” the subject “Alice” provides information about who performed the action, but occupies different relative positions in the two windows. A content-dependent mixer can adjust its weights to emphasize this information as its position changes, whereas a fixed convolution applies the same coefficients at each relative position. The construction of recurrent training pairs therefore lacks an explicit mechanism for adapting local aggregation weights to the information present in each window. This motivates content-dependent local selection before recurrent learning.

\paragraph{Model variants.}
We obtain SWGDN2 and SWKDA by replacing the depthwise short convolutions
on Q, K, and V in GDN2 and KDA, respectively, with sliding-window
attention. The suffix \texttt{-G}\(n\) denotes \(n\)
channel groups per head in each branch.

\subsection{Sliding-Window Attention for Local Mixing}
\label{sec:pseudo_query_swa}

Let $d$ be the head dimension, and partition the channels of each head
into $G$ groups. Each branch $X \in \{Q,K,V\}$ learns a separate
$G \times d$ tensor per head. Its row $p^{X,g}$ is the learned
$d$-dimensional pseudo-query for group $g$.

At time step $t$, the causal window is
$\mathcal{W}_t = \{\max(1,t-W+1),\ldots,t\}$.
 We set $W=4$ to match the kernel
size of the original short convolution.

In the recurrent update, keys serve as regression inputs and determine
the directions along which the state is updated. We therefore reuse
the projected keys before local mixing to represent candidate tokens
for scoring in all three branches. Each branch scores the candidates
with its own learned queries and uses the resulting weights to mix
its projected queries, keys, or values.

We apply RMS normalization~\citep{zhang2019rmsnorm} followed by
rotary position embeddings (RoPE)~\citep{su2024roformer} to the
learned queries and projected keys. RoPE makes the scores sensitive
to relative offsets within the window. Denoting the transformed
vectors by $p_t^{X,g}$ and $k_i$, respectively, we compute
the mixing weights as
\begin{equation}
a^{X,g}_{t,i}
=
\operatorname{softmax}_{i \in \mathcal{W}_t}
\left(
\frac{(p_t^{X,g})^\top k_i}{\sqrt{d}}
\right).
\end{equation}

Within each branch and group, the same learned query is reused
as the window advances, while the scores depend on the candidate keys
and their relative offsets. A source token can therefore receive high
weight in successive windows, with its weight at each step also
depending on the other candidates in the window.

For branch $X$, let $x_i^g$ denote the projected channels belonging to group $g$ at position $i$. The local aggregation for this group is defined as
\begin{equation}
\widetilde{x}_t^g
=
\operatorname{SiLU}
\left(
\sum_{i \in \mathcal{W}_t}
a^{X,g}_{t,i} x_i^g
\right),
\qquad X \in \{Q,K,V\}.
\end{equation}

Each group shares a single set of attention weights across its $d/G$ channels. The aggregated outputs of all groups are concatenated to obtain the locally mixed Q, K, and V representations at the current position.When $G=1$, all channels within a head share the same attention weights, whereas $G=d$ assigns an independent set of attention weights to each channel. The resulting Q, K, and V representations are subsequently fed into the original recurrent computation. Importantly, the state update rules and chunkwise parallel algorithms of GDN2 and KDA remain unchanged.

\section{How to understand SWGDN}

To understand how sliding-window attention affects local mixing and the key–value pairs used in recurrent updates, we conduct a series of diagnostic experiments, analyze the resulting patterns, and offer possible explanations.

\subsection{What Sliding Window Attention Learned?}

We analyze the local mixing weights from the final SWKDA and SWGDN2 checkpoints trained at the 340M/15B scale, using 160 validation sequences of 2048 tokens each and collecting statistics across all 18 layers. The window size is set to 4, and only complete windows are considered. The results are first averaged across heads and then equally averaged across layers.
\begin{table}[htbp]
\centering
\caption{Local attention statistics for G1 models at 340M/15B.}
\label{tab:local_attention}
\small
\setlength{\tabcolsep}{3pt}
\begin{tabular}{@{}llcccc@{}}
\toprule
Model & Branch & Entropy & Max. & Current & Previous \\
\midrule
SWKDA-G1 & Q & 0.568 & 0.800 & 0.94 & 0.06 \\
 & K & 0.658 & 0.742 & 0.62 & 0.32 \\
 & V & 0.608 & 0.764 & 0.61 & 0.35 \\
\midrule
SWGDN2-G1 & Q & 0.616 & 0.767 & 0.70 & 0.29 \\
 & K & 0.712 & 0.715 & 0.67 & 0.27 \\
 & V & 0.730 & 0.705 & 0.52 & 0.42 \\
\bottomrule
\end{tabular}
\end{table}

Both models assign substantial attention mass to a few offsets (\cref{tab:local_attention}). In SWKDA-G1, the Q branch most often selects the current position, while K and V select the previous position more frequently; in SWGDN2-G1, the V branch selects the previous position in 42\% of windows. These patterns suggest that the readout query remains largely anchored to the current token, while K/V mixing can carry information from the same source token into successive writes as the window advances.

\textbf{Similarity Between Adjacent Learning Pairs.}
We compare KDA and GDN2 with their respective G1 variants at the final 340M/15B checkpoints. All four models use the same 32 frozen validation sequences and predefined anchors. Each sequence has 2048 tokens and 16 predefined anchors. We report results across all 16 heads in layers 1, 9, and 18. We measure keys after local mixing and L2 normalization and values after local mixing and activation, before the write gate.

We measure the difference between vectors using a symmetric relative distance and define the joint distance between learning pairs as the maximum of the K and V distances:
\[
\begin{aligned}
d(x,y)
&=\frac{2\lVert x-y\rVert_2}
{\lVert x\rVert_2+\lVert y\rVert_2+\epsilon},\\
D_{KV}(t,\tau)
&=\max\left(
d(k_t,k_{t+\tau}),
d(v_t,v_{t+\tau})
\right).
\end{aligned}
\]

We average the joint distances within each layer and offset, excluding pairs with near-zero vectors.

\begin{table}[htbp]
\centering
\caption{Mean joint K/V distances at offsets of 1--3 tokens. Lower distances indicate greater similarity. }
\label{tab:kv_similarity}
\small
\setlength{\tabcolsep}{5pt}
\begin{tabular}{@{}lrrr@{}}
\toprule
Layer & Offset 1 & Offset 2 & Offset 3 \\
\midrule
\multicolumn{4}{@{}l}{\textit{KDA / SWKDA-G1}} \\
1 & 1.269 / \textbf{0.995} & 1.258 / \textbf{1.069} & 1.249 / \textbf{1.058} \\
9 & 1.081 / \textbf{0.773} & 1.227 / \textbf{0.997} & 1.268 / \textbf{1.065} \\
18 & 1.237 / \textbf{0.689} & 1.245 / \textbf{0.842} & 1.240 / \textbf{0.869} \\
\midrule
\multicolumn{4}{@{}l}{\textit{GDN2 / SWGDN2-G1}} \\
1 & 1.168 / \textbf{0.677} & 1.168 / \textbf{0.946} & 1.142 / \textbf{1.004} \\
9 & 1.128 / \textbf{0.770} & 1.224 / \textbf{1.022} & 1.235 / \textbf{1.087} \\
18 & 0.998 / \textbf{0.778} & 1.075 / \textbf{0.962} & 1.089 / \textbf{1.018} \\
\bottomrule
\end{tabular}
\end{table}

Both G1 variants yield lower mean joint K/V distances than their respective short-convolution baselines at every measured layer and offset. When both the key and value mixers consistently assign high weight to the same source token across successive windows, the resulting regression examples can be close to a common key–value pair. If decay and gating have limited effects, successive updates can then be interpreted as taking additional stochastic gradient steps to refine the fit to that pair.

\subsection{Local Selection and Approximate Multi-Step Learning}
\label{sec:local_selection}

From the TTR perspective, similar mixed keys give successive writes similar key-side update directions, and similar mixed values give them similar regression targets before gating. Consider two successive updates whose learning pairs are close to a common pair.
\begin{equation}
(\widetilde{k}_t,\widetilde{v}_t)
\approx
(\widetilde{k}_{t+1},\widetilde{v}_{t+1})
\approx
(k^\star,v^\star).
\end{equation}

Ignoring decay and gating, successive Delta Rule updates on these pairs approximate multiple gradient descent steps on the same regression objective.
\begin{equation}
\ell^\star(S)
=
\frac{1}{2}
\left\|Sk^\star-v^\star\right\|_2^2.
\end{equation}

This approximation is most accurate when the mixed pairs are close and decay and gating have little effect on the updates. In GDN2, the erase and write gates can change the effective learning pair even when the mixed keys and values are similar. \Cref{sec:attention_analysis,sec:gate_discussion} detail the effects of decay and gating.

\section{Experiment}

\begin{table*}[t]
\centering
\caption{Language modeling and zero-shot reasoning performance of 340M-parameter models trained on 15B tokens from FineWeb.}
\label{tab:lm_reasoning}
\setlength{\tabcolsep}{4pt}
\resizebox{\textwidth}{!}{%
\begin{tabular}{lc|ccc|cccccccc|c}
\toprule
Model & FineWeb ppl & Wiki. & LMB. & LMB. & PIQA & Hella. & Wino. & ARC-e & ARC-c & OBQA & SIQA & BoolQ & Avg. \\
 & $\downarrow$ & ppl$\downarrow$ & ppl$\downarrow$ & acc$\uparrow$ & acc$\uparrow$ & acc$\uparrow$ & acc$\uparrow$ & acc$\uparrow$ & acc$\uparrow$ & acc$\uparrow$ & acc$\uparrow$ & acc$\uparrow$ & acc$\uparrow$ \\
\midrule
GDN2 & 11.6180 & 29.6074 & 32.1497 & 33.55 & 66.05 & 39.20 & \textbf{53.51} & 56.82 & 25.26 & 22.60 & 39.25 & 54.80 & 44.69 \\
GDN2 with DynamicConv & 11.4713 & 28.9051 & 32.2360 & 33.11 & 66.38 & 39.80 & 52.57 & 59.05 & 26.28 & 22.40 & \textbf{39.46} & 50.24 & 44.52 \\
SWGDN2-G1 & 11.4657 & 28.8916 & 29.9178 & 33.94 & 65.51 & 39.90 & 52.88 & 57.07 & 25.26 & 22.80 & 38.74 & 57.86 & 45.00 \\
SWGDN2-G16 & 11.4626 & 28.9036 & 30.4465 & 35.05 & 66.43 & 39.80 & 51.22 & 58.96 & 23.63 & 21.40 & 39.00 & 58.96 & 44.93 \\
SWGDN2-G64 & 11.4473 & 29.1324 & 31.9430 & 33.96 & 65.72 & 39.63 & 53.04 & 56.65 & 25.51 & 22.80 & 38.79 & \textbf{59.05} & 45.15 \\
KDA & 11.1157 & 27.4184 & 28.8534 & 34.50 & \textbf{67.41} & 41.60 & 52.49 & 60.52 & 26.11 & \textbf{23.40} & 38.64 & 54.83 & 45.62 \\
SWKDA-G1 & \textbf{10.9243} & \textbf{27.0141} & \textbf{27.2252} & \textbf{36.77} & 67.19 & \textbf{41.81} & 53.35 & \textbf{61.03} & \textbf{26.96} & 23.20 & 37.41 & 57.92 & \textbf{46.11} \\
\bottomrule
\end{tabular}}
\par\vspace{0.8em}
\centering
\caption{RULER needle-in-a-haystack retrieval accuracy (\%). S-NIAH-3 uses UUID targets and MK-NIAH-1 uses four keys.}
\label{tab:niah}
\setlength{\tabcolsep}{3pt}
\resizebox{\textwidth}{!}{%
\begin{tabular}{lcccccccccccccc|c}
\toprule
& \multicolumn{4}{c}{S-NIAH-1} & \multicolumn{4}{c}{S-NIAH-2} & \multicolumn{3}{c}{S-NIAH-3} & \multicolumn{3}{c}{MK-NIAH-1} & \\
\cmidrule(lr){2-5}\cmidrule(lr){6-9}\cmidrule(lr){10-12}\cmidrule(lr){13-15}
Model & 1K & 2K & 4K & 8K & 1K & 2K & 4K & 8K & 1K & 2K & 4K & 1K & 2K & 4K & Avg. \\
\midrule
GDN2 & 99.4 & 94.4 & 56.8 & 18.0 & 62.0 & 47.8 & 27.2 & 9.6 & 6.0 & 5.6 & 3.0 & 16.8 & 19.2 & \textbf{24.0} & 35.0 \\
GDN2 + Dyn & \textbf{100.0} & 93.6 & 37.4 & 15.4 & 49.6 & 64.8 & 23.2 & 12.0 & 1.0 & 1.0 & 0.0 & \textbf{24.8} & \textbf{22.6} & 15.2 & 32.9 \\
SWGDN2-G1 & 99.8 & 98.4 & 57.6 & 19.6 & 99.4 & 86.2 & 25.8 & 10.4 & 9.4 & 1.6 & 0.6 & 18.8 & 21.2 & 15.6 & 40.3 \\
SWGDN2-G16 & 98.4 & 72.0 & 22.6 & 12.2 & 98.6 & 87.2 & \textbf{50.2} & 13.2 & 6.8 & 2.6 & 0.8 & 21.4 & 22.2 & 23.6 & 38.0 \\
SWGDN2-G64 & 99.6 & 98.6 & \textbf{87.8} & \textbf{38.6} & \textbf{99.6} & 83.8 & 33.6 & 7.2 & 0.0 & 0.8 & 0.4 & 22.8 & 21.2 & 20.6 & \textbf{43.9} \\
KDA & \textbf{100.0} & 98.8 & 69.2 & 17.8 & 99.2 & 61.0 & 16.4 & 6.8 & 14.4 & 2.0 & \textbf{5.2} & 20.6 & 15.0 & 17.0 & 38.8 \\
SWKDA-G1 & 99.8 & \textbf{99.0} & 73.4 & 20.4 & 98.4 & \textbf{94.4} & 16.2 & \textbf{15.4} & \textbf{16.8} & \textbf{13.6} & 0.6 & 22.2 & 20.6 & 20.2 & 43.6 \\
\bottomrule
\end{tabular}}
\end{table*}

\paragraph{Setup.} We compare seven approximately 340M-parameter models across the GDN2 and KDA backbones~\cite{hatamizadeh_gated_2026,kimiteam2025kimilinearexpressiveefficient}. The KDA and GDN2 baselines use width-4 short convolutions on Q/K/V. We evaluate GDN2 with DynamicConv~\cite{sieberling2026dynamicshortconvolutionsimprove} , SWGDN2 with G1, G16, and G64 grouping and SWKDA with G1 grouping. All models have 18 layers, hidden size 1024, and 16 heads of dimension 64; the MLP width is 2816 for GDN2 and 3488 for KDA. Each model is trained on the same fixed sequence of FineWeb-Edu windows~\cite{penedo2024finewebdatasetsdecantingweb} for approximately 15B tokens, with a sequence length of 2048. See \cref{app:experimental_config,app:local_mixing} for training details and the mixer comparison.

\paragraph{Language modeling and zero-shot downstream tasks.} \Cref{tab:lm_reasoning} reports WikiText-2 perplexity~\cite{merity2016pointersentinelmixturemodels}, and LAMBADA perplexity~\cite{swgdn_exp_paperno2016}. We also evaluate eight zero-shot tasks\cite{bisk2019piqareasoningphysicalcommonsense,zellers2019hellaswagmachinereallyfinish,sakaguchi2019winograndeadversarialwinogradschema,clark2018thinksolvedquestionanswering,mihaylov2018suitarmorconductelectricity,sap2019socialiqacommonsensereasoningsocial,clark2019boolqexploringsurprisingdifficulty}, reporting length-normalized accuracy for HellaSwag and accuracy for the other tasks. All sliding-window variants reduce perplexity on the language-modeling measures relative to their respective baselines. 

\paragraph{Synthetic needle-in-a-haystack tasks.} \Cref{tab:niah} reports four needle-in-a-haystack variants from RULER~\cite{swgdn_exp_hsieh2024} at context lengths of 1K--8K, depending on the task. All SWGDN variants outperform their respective baselines on average across the 14 RULER settings. SWGDN2-G1 improves S-NIAH-2 accuracy at 1K from 62.0\% to 99.4\% over GDN2, while SWGDN2-G64 improves S-NIAH-1 at 4K from 56.8\% to 87.8\%. All sliding-window attention variants achieved better average performance than the baseline.

\paragraph{Ablation study.} \Cref{tab:local_mixing_ablation} evaluates the effects of applying SWA to different branches in SWGDN, as well as the impact of incorporating RoPE into SWA. We observe several interesting phenomena.First, applying SWA exclusively to the K and V branches yields better performance on the Needle-in-a-Haystack (NIAH) task than the original SWGDN2, whereas applying SWA only to the Q branch leads to a performance degradation. We hypothesize that this phenomenon can be attributed to two factors. First, our SWA mechanism reuses the projected keys to compute attention scores. However, keys and queries serve distinct roles in linear attention. Consequently, using keys to determine the local mixing weights of the Q branch may introduce undesirable interference. One potential direction for future work is to compute attention scores independently using the corresponding projections of each branch.Second, our analysis of the attention weight distributions in SWGDN reveals that the Q branch assigns the largest attention weight to the current position in most cases. Therefore, removing local mixing from the Q branch has a relatively limited impact on model performance, whereas removing local mixing from the K and V branches leads to performance degradation.This pattern is consistent with our earlier intuition that, from a TTR perspective, how key–value pairs are constructed for recurrent updates can affect how well past associations are retrieved.

Furthermore, we observe that removing RoPE from SWA substantially improves performance on the NIAH task. We hypothesize that this improvement arises because the relative positional shifts induced by the sliding window affect attention scores through RoPE. Consider two identical tokens within a window. Without positional encoding (NoPE), their relative attention weights remain unchanged solely with respect to the movement of the window. In contrast, RoPE introduces position-dependent variations in attention scores, potentially reducing the ability of an important token to maintain consistently high attention weights across successive time steps.

\begin{table}[tb]
\centering
\caption{Ablation results for SWGDN2. Zero-shot and RULER NIAH accuracy average. Branches without SWA use width-4 short convolutions.}
\label{tab:local_mixing_ablation}
\footnotesize
\setlength{\tabcolsep}{2pt}
\begin{tabular*}{\columnwidth}{@{\extracolsep{\fill}}lcrrr@{}}
\toprule
SWA branches & RoPE & \shortstack{FineWeb \\ppl $\downarrow$} & \shortstack{Zero-shot\\(\%) $\uparrow$} & \shortstack{RULER NIAH\\(\%) $\uparrow$} \\
\midrule
K, V & Yes & 11.5208 & 44.73 & \textbf{44.9} \\
Q & Yes & 11.5613 & 44.28 & 29.4 \\
Q, K, V & No & 11.7357 & 44.83 & 44.7 \\
\bottomrule
\end{tabular*}
\end{table}

\paragraph{Generative recall tasks.} \Cref{tab:additional_tasks} reports six recall-intensive tasks: SWDE, SQuAD-completion, FDA, TriviaQA, Natural Questions, and DROP. The evaluation follows the recall-task setting studied by \citet{swgdn_exp_arora2024}. G1 increases the average from 19.36\% to 20.23\% on GDN2 and from 19.91\% to 21.63\% on KDA. Compared with KDA, SWKDA-G1 gains 4.41 and 4.62 percentage points on SWDE and FDA, respectively.
\begin{table}[!tb]
\centering
\caption{Accuracy on real-world retrieval tasks. Dyn. denotes DynamicConv. G1, G16, and G64 are sliding-window variants of the indicated backbone.}
\label{tab:additional_tasks}
\footnotesize
\setlength{\tabcolsep}{2pt}
\begin{tabular*}{\columnwidth}{@{\extracolsep{\fill}}lrrrrrrr@{}}
\toprule
 & \multicolumn{5}{c}{GDN2 backbone} & \multicolumn{2}{c@{}}{KDA backbone} \\
\cmidrule(lr){2-6}\cmidrule(l){7-8}
Task & GDN2 & +Dyn. & G1 & G16 & G64 & KDA & G1 \\
\midrule
SWDE & 16.92 & 15.03 & 16.29 & \underline{17.19} & 15.48 & 15.66 & \textbf{20.07} \\
SQuAD & 24.66 & 24.20 & \textbf{26.64} & 23.59 & 25.03 & 26.07 & \underline{26.41} \\
FDA & \underline{3.18} & 2.45 & \underline{3.18} & 2.27 & 2.27 & \underline{3.18} & \textbf{7.80} \\
TQA & 44.91 & 46.45 & 47.45 & 46.03 & 47.16 & \underline{47.51} & \textbf{48.76} \\
NQ & 8.08 & \underline{8.65} & 8.49 & \textbf{8.90} & 8.20 & 8.55 & \textbf{8.90} \\
DROP & 18.40 & 18.35 & \textbf{19.31} & 16.29 & \underline{18.88} & 18.50 & 17.82 \\
\midrule
Avg. & 19.36 & 19.19 & \underline{20.23} & 19.05 & 19.50 & 19.91 & \textbf{21.63} \\
\bottomrule
\end{tabular*}
\end{table}

\begin{figure*}[!t]
\centering
\begin{minipage}[t]{0.48\textwidth}
\centering
\includegraphics[width=\textwidth]{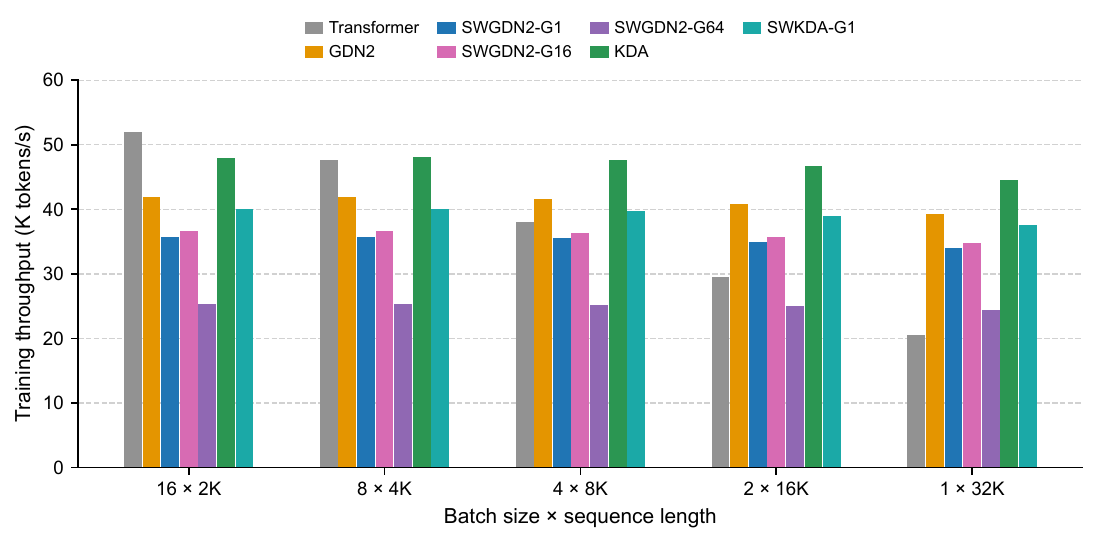}
\caption{Training throughput on a NVIDIA RTX 6000D GPU.}
\label{fig:training_throughput}
\end{minipage}\hfill
\begin{minipage}[t]{0.48\textwidth}
\centering
\includegraphics[width=\textwidth]{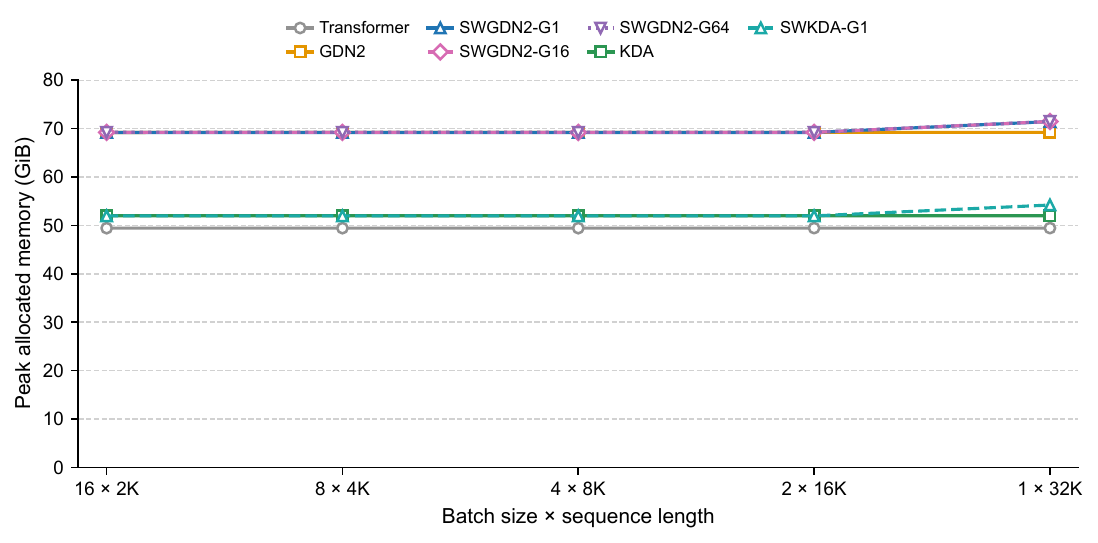}
\caption{Peak Allocated Memory.}
\label{fig:training_memory}
\end{minipage}
\end{figure*}

\paragraph{Training efficiency.} We benchmark full training steps on one NVIDIA RTX 6000D. Each step includes forward and backward passes and an optimizer update. \Cref{fig:training_throughput,fig:training_memory} report throughput and peak allocated memory, respectively. At batch size 16 and length 2K, SWGDN2-G1 and SWKDA-G1 attain 85.3\% and 83.5\% of their respective baselines' throughput with similar peak allocations. At batch size 1 and length 32K, their peak allocations increase by approximately 2.24 GiB relative to the respective baselines.

\section{Related Work}

\paragraph{Multi-sample memory updates.} From a test-time regression perspective, recurrent memory updates can be interpreted as learning key--value associations from context. DeltaProduct~\cite{siems2025deltaproduct} extends DeltaNet by constructing multiple key--value pairs for each token and applying multiple sequential updates, increasing the expressivity of the state transition. ATLAS~\cite{behrouz2025atlaslearningoptimallymemorize} takes a different approach: its $\Omega$-rule incorporates a weighted sum of losses over a local context into the memory optimization objective. In contrast, SWGDN mixes local representations before constructing the learning pair used by each time step, while retaining the original recurrent update.Due to the overlap between adjacent windows and the dynamic selectivity of SWA, the key–value (KV) pair associated with a given token can receive high attention weights across multiple time steps, which makes the resulting KV pairs more likely to exhibit similarity across subsequent time steps. These updates can be interpreted as learning from related examples, subject to the effects of gating and decay.

\paragraph{Combining attention and linear recurrence.} 
Hybrid architectures combine different sequence mixers across layers or within individual layers~\cite{ren2025sambasimplehybridstate,lieber2024jambahybridtransformermambalanguage}. In inter-layer hybrids, Gemma 2\cite{gemmateam2024gemma2improvingopen} alternates sliding-window attention (SWA) with global attention, and Kimi Linear~\cite{kimiteam2025kimilinearexpressiveefficient} inserts one global multi-head latent attention layer after every three KDA layers. In intra-layer hybrids, Routing Transformer~\cite{roy2020efficientcontentbasedsparseattention} assigns different heads in the same layer to local attention and content-based routing attention, and Native Sparse Attention(NSA)~\cite{yuan2025nativesparseattentionhardwarealigned} uses learned gates to fuse the outputs of its compression, sparse selection, and sliding-window branches

SWGDN offers another way to combine these mechanisms by embedding SWA directly inside the linear attention. SWA replaces the short convolution and performs content-dependent local mixing of the projected queries, keys, and values. The mixed keys and values drive the recurrent state update, and the mixed queries read from that state. Local attention thus shapes both what is written to memory and how it is read. The resulting operator occupies a middle ground between SWA and linear attention, combining explicit token mixing within a local window with recurrent propagation of the aggregated information beyond it.

\section{Limitations}
Our main experiments use 340M-parameter models trained on 15B tokens. Some ablated configurations improve needle-in-a-haystack retrieval relative to the full model, suggesting that the choice of local mixing design warrants further study. Our compute budget limited the range of designs and model scales we could evaluate. 

The computational cost of local attention limits the practicality of using separate attention weights for every channel. A depthwise short convolution applies a learned filter independently to each channel, whereas in SWGDN each attention group scores the tokens in its window using full-dimensional keys. Assigning one group to every channel therefore makes the per-head scoring cost grow quadratically with head dimension at a fixed window width. We evaluated this per-channel configuration only at a head dimension of 64 (G64), where it already incurred a substantial training throughput penalty relative to GDN2 with depthwise short convolutions (Figure 2)

\section{Conclusion}
We presented SWGDN, which integrates local softmax attention into linear recurrent models by mixing queries, keys, and values before they are used for state readout and updates. This design lets content-dependent local selection shape both how the recurrent state is updated and how it is read, while preserving the original recurrence. In 340M GDN2 and KDA models, SWGDN lowers perplexity and improves average performance on generative recall tasks. These results motivate a broader question for hybrid attention design: Could refining local representations before recurrent processing offer another way to improve linear attention?


\bibliographystyle{icml2027}
\bibliography{references}


\clearpage
\appendix
\onecolumn
\raggedbottom
\makeatletter
\setlength{\@fptop}{0pt}
\setlength{\@fpsep}{12pt}
\setlength{\@fpbot}{0pt plus 1fil}
\makeatother

\section{Linear Attention}
\label{app:linear_attention_derivation}

This section outlines the development of recurrent linear attention from additive state updates to the gated delta rules used by our backbones. These models address related questions about memory and efficient computation, but do not form a single chronological chain of direct successors. We use the main-text convention $S_t\in\mathbb{R}^{d_v\times d_k}$, with column-vector queries and keys in $\mathbb{R}^{d_k}$ and values in $\mathbb{R}^{d_v}$.

\subsection{From kernel attention to a recurrent state}
\label{app:kernel_attention}

In causal softmax attention, each query reads a weighted combination of the values at all preceding positions. Explicitly evaluating all query--key interactions requires quadratic work in sequence length. Kernel linear attention instead uses a similarity of the form $\kappa(q_t^0,k_i^0)=\phi(q_t^0)^\top\phi(k_i^0)$, where $\phi$ maps inputs to a finite-dimensional feature space. Writing $q_t=\phi(q_t^0)$ and $k_t=\phi(k_t^0)$, associativity gives~\cite{katharopoulos_transformers_2020}
\begin{align*}
o_t &= \frac{\sum_{i=1}^{t}v_i\,\kappa(q_t^0,k_i^0)}{\sum_{i=1}^{t}\kappa(q_t^0,k_i^0)}
 = \frac{C_tq_t}{n_t^\top q_t},\\
C_t &= C_{t-1}+v_tk_t^\top,\qquad
n_t=n_{t-1}+k_t,\qquad C_0=0,\quad n_0=0.
\end{align*}
The expression assumes a nonzero denominator. With fixed feature dimensions, the state occupies $O(d_vd_k+d_k)$ space and each recurrent update and readout takes $O(d_vd_k)$ work, independent of prefix length. These statements concern dependence on sequence length; state size and work still depend on the feature and value dimensions. The feature map determines the kernel, and a finite-dimensional feature map is not in general an exact representation of the softmax kernel\cite{
  arora2024zoology,
  arora2024based,
  jelassi2024repeat,
  wen2025rnns}.

\subsection{Additive memory and fast weights}

In the unnormalized form, $C_t$ is a matrix-valued memory read by $o_t=C_tq_t$. Each outer product $v_tk_t^\top$ adds an association between a key and a value, so a later query retrieves a superposition of the stored values weighted by its similarity to their keys. This connects linear attention to fast-weight memory, whose state changes with the input sequence while the parameters that produce the keys and values are learned across training examples~\cite{schlag_linear_2021}.

The memory has a fixed size, however, and different associations share its entries. A write can therefore affect the readout of other, nonorthogonal keys. In addition, the additive rule writes $v_tk_t^\top$ without checking what the existing state already predicts for $k_t$. Repeated or conflicting associations can accumulate rather than explicitly correct an earlier mapping. These observations motivate both mechanisms that control state retention and updates that use the current memory prediction.

\subsection{Forgetting and data-dependent gating}

One way to control the influence of earlier tokens is to decay the state before adding the next association. In a scalar-decay example, $S_t=\alpha_t S_{t-1}+v_tk_t^\top$. Retentive networks use retention with fixed decay factors across different timescales, while gated linear attention introduces input-dependent gates and a hardware-efficient chunkwise implementation~\cite{sun2023retentivenetworksuccessortransformer,yang2024gatedlinearattentiontransformers}. Finer-grained gates can control retention along state dimensions. RWKV and selective state-space models, including Mamba, are related recurrent sequence-modeling approaches that also use mechanisms for controlling state evolution, although their parameterizations and architectures differ~\cite{peng2023rwkvreinventingrnnstransformer,gu2024mambalineartimesequencemodeling}.

Decay controls how much old state is retained before a new write. By itself, it does not use the current key to calculate and correct the state's prediction for that association. The delta rule provides a complementary mechanism for such a correction.

\subsection{DeltaNet and test-time regression}
\label{app:delta_gradient}

The delta rule treats the state as a fast regression model: a key is an input, its value is a target, and a query reads the resulting model~\cite{schlag_linear_2021,wang_test-time_2026}. For the current pair, define $\ell_t(S)=\frac12\lVert Sk_t-v_t\rVert_2^2$. A gradient step from the preceding state gives~\cite{yang_parallelizing_2025}
\begin{align*}
S_t &= S_{t-1}-\beta_t\nabla_S\ell_t(S_{t-1})\\
&= S_{t-1}+\beta_t(v_t-S_{t-1}k_t)k_t^\top .
\end{align*}
The state first predicts the value at $k_t$ and then writes a correction proportional to the residual. For a unit-norm key, the residual on this pair is multiplied by $1-\beta_t$; with $\beta_t=1$, the immediate post-update state satisfies $S_tk_t=v_t$. Subsequent updates can still change this association when their keys are not orthogonal to $k_t$.

The regression perspective also clarifies the difference from additive memory. For the unnormalized prefix loss $\mathcal L_t(S)=\frac12\sum_{i=1}^t\lVert Sk_i-v_i\rVert_2^2$, one unit-size batch gradient step from $S=0$ gives $\sum_{i=1}^t v_i k_i^\top$. DeltaNet instead carries its state forward and updates it sequentially using the current residual. Other approaches modify the regression procedure; for example, PGDN uses a history-dependent preconditioner to adjust the update direction~\cite{tumma_preconditioned_2026}.

\subsection{Gated DeltaNet and Kimi Delta Attention}

Gated DeltaNet combines state decay with a delta update, allowing the model to control retention and correct the current association in the same recurrence~\cite{yang2025gateddeltanetworksimproving}. With scalar decay $\alpha_t$, its update can be written as
\begin{align*}
\bar S_t &= \alpha_t S_{t-1},\\
S_t &= \bar S_t+\beta_t(v_t-\bar S_tk_t)k_t^\top.
\end{align*}
The residual is evaluated against the decayed state. Thus decay changes the memory presented to the correction step, while the delta term adjusts the mapping along the current key direction.

Kimi Delta Attention (KDA), a module in the Kimi Linear architecture, extends this design with channel-wise decay on the key dimension~\cite{kimiteam2025kimilinearexpressiveefficient}. In our convention, $\bar S_t=S_{t-1}\operatorname{Diag}(\alpha_t)$, where $\alpha_t\in\mathbb R^{d_k}$, followed by the same scalar-gated delta update above. The vector decay permits different key-space coordinates to retain different amounts of the previous state. KDA and Gated DeltaNet both couple the strengths of the correction's erase and write components through the scalar $\beta_t$.

\subsection{GDN2: decoupled erase and write}

Gated DeltaNet-2 (GDN2) retains channel-wise decay while separating the controls for reading old content and writing a new value~\cite{hatamizadeh_gated_2026}. Let $b_t\in\mathbb R^{d_k}$ be the erase gate and $w_t\in\mathbb R^{d_v}$ the write gate. Its update is
\begin{align*}
\bar S_t &= S_{t-1}\operatorname{Diag}(\alpha_t),\\
e_t &= b_t\odot k_t,\qquad z_t=w_t\odot v_t,\\
S_t &= \bar S_t+(z_t-\bar S_te_t)k_t^\top.
\end{align*}
The erase gate changes the key coordinates used to read the old state, whereas the write gate changes the value components supplied to the update. The outer-product direction remains $k_t$. If both gates are tied to the same scalar $\beta_t$, the update reduces to the KDA form. For positive erase gates, the update also admits a preconditioned-gradient interpretation under a gated regression loss, as noted in the main text; this interpretation does not change the native recurrence.

\subsection{Similar Learning Pairs and Successive Updates}
\label{sec:attention_analysis}

KDA employs channel-wise decay and a scalar update rate. Its state update can be written as
\[
\bar{S}_t = S_{t-1}D_t,\qquad
S_t = \bar{S}_t + \beta_t(v_t-\bar{S}_tk_t)k_t^\top,
\]
where \(D_t\) is a diagonal matrix representing channel-wise decay, and \(\beta_t\) denotes the update rate. Let \(r_t=v_t-\bar{S}_tk_t\) denote the pre-update residual for the current learning pair. The post-update residual for the same learning pair is then
\[
\begin{aligned}
r_t^{\mathrm{post}}
&=v_t-S_tk_t\\
&=(1-\beta_t\lVert k_t\rVert_2^2)r_t.
\end{aligned}
\]

When \(0<\beta_t\lVert k_t\rVert_2^2<2\), this update reduces the residual magnitude for the current learning pair. If subsequent updates use the same learning pair and decay is temporarily ignored, successive updates correspond to performing multiple gradient descent steps on the same regression objective.

In practice, both the learning pairs and decay factors vary across time steps. Comparing the pre-update residual at the next step with the post-update residual at the current step yields

\[
\begin{aligned}
r_{t+1}
&=v_{t+1}-S_tD_{t+1}k_{t+1}\\
&=r_t^{\mathrm{post}}+(v_{t+1}-v_t)\\
&\quad-S_t(D_{t+1}k_{t+1}-k_t).
\end{aligned}
\]

The second term captures changes in the target, while the third term accounts for the effects of key variations and decay on the prediction. If these deviations are small relative to the residual, the residual encountered by the next update remains close to that left by the previous update. In this case, successive updates can be approximately interpreted as multi-step learning toward related targets.

\subsection{Decoupled Erase and Write Gates}
\label{sec:gate_discussion}

To see how the gates affect an earlier association, consider the effective pair $(e_t,z_t)$ defined for GDN2 in the background section. For KDA, take $e_t=\beta_t k_t$ and $z_t=\beta_t v_t$. A later write at $u>t$ changes the prediction at $e_t$ by
\[
(S_u-\bar S_u)e_t
=(k_u^\top e_t)(z_u-\bar S_u e_u).
\]
The effect on the old association depends on the key overlap and on the direction and magnitude of this change relative to the post-decay anchor residual. In KDA, a single $\beta_u$ scales both the target and the read from the decayed state, so it scales the correction for fixed $k_u$, $v_u$, and $\bar S_u$. In GDN2, $b_u$ changes the key-side read and $w_u$ independently changes the value-side target. The resulting correction can change direction even when adjacent ungated K/V pairs are similar~\cite{kimiteam2025kimilinearexpressiveefficient,hatamizadeh_gated_2026}.

\section{Experimental Configuration}
\label{app:experimental_config}

\paragraph{Architecture.} \Cref{tab:appendix_model_sizes} reports the architecture and exact parameter count of each 340M model. All models use a vocabulary of 32,000 tokens and a causal width-4 mixer on projected Q, K, and V. The SW variants use 1, 16, or 64 groups per head and branch, corresponding to 64, 4, or 1 payload channels per group. The DynamicConv implementation uses a rank-26 kernel generator for each branch.

\begin{table}[htbp]
\centering
\caption{Model configurations for the 340M experiments. The key and value dimensions $d_k$ and $d_v$ are per head. State size is the recurrent state shape $H\times d_v\times d_k$ per layer and sequence, where $H$ is the head count.}
\label{tab:appendix_model_sizes}
\small
\setlength{\tabcolsep}{3.5pt}
\begin{tabular}{@{}lrrrrrrllr@{}}
\toprule
Model & Depth & Heads & $d_k$ & $d_v$ & $d_{\mathrm{model}}$ & $d_{\mathrm{MLP}}$ & State size & Local mixer & Parameters \\
\midrule
GDN2 & 18 & 16 & 64 & 64 & 1,024 & 2,816 & $16\times64\times64$ & Depthwise conv. & 339,511,712 \\
GDN2 with DynamicConv & 18 & 16 & 64 & 64 & 1,024 & 2,816 & $16\times64\times64$ & DynamicConv ($R=26$) & 346,700,192 \\
SWGDN2-G1 & 18 & 16 & 64 & 64 & 1,024 & 2,816 & $16\times64\times64$ & SWA $G=1$ & 339,345,824 \\
SWGDN2-G16 & 18 & 16 & 64 & 64 & 1,024 & 2,816 & $16\times64\times64$ & SWA $G=16$ & 340,175,264 \\
SWGDN2-G64 & 18 & 16 & 64 & 64 & 1,024 & 2,816 & $16\times64\times64$ & SWA $G=64$ & 342,829,472 \\
KDA & 18 & 16 & 64 & 64 & 1,024 & 3,488 & $16\times64\times64$ & Depthwise conv. & 339,216,800 \\
SWKDA-G1 & 18 & 16 & 64 & 64 & 1,024 & 3,488 & $16\times64\times64$ & SWA $G=1$ & 339,050,912 \\
\bottomrule
\end{tabular}
\end{table}

\paragraph{Data and optimization.} We train all seven 340M models on FineWeb-Edu for 57,221 optimizer steps (approximately 15B tokens), with 128 sequences of 2,048 tokens per step. We use AdamW with $(\beta_1,\beta_2)=(0.9,0.95)$, weight decay 0.1, gradient-norm clipping at 1.0, and a peak learning rate of $4\times10^{-4}$. We linearly warm up for the first 1\% of steps and then use cosine decay to $4\times10^{-5}$. Training uses BF16 autocast with FP32 parameters.

\section{A Banded-Matrix View of Local Mixing}
\label{app:local_mixing}

\subsection{The window defines the matrix support}

Let $x_{t,c}$ be the projected payload in channel $c$ of one Q, K, or V branch, and collect its values over a sequence of length $T$ in $\mathbf{x}_c$. Before the local activation, mixing along time can be written as
\[
\mathbf{m}_c=M^{(c)}\mathbf{x}_c,\qquad
m_{t,c}=\sum_{j=\max(1,t-W+1)}^t M^{(c)}_{t,j}\,x_{j,c}.
\]
Row $t$ collects the coefficients used at output position $t$, and column $j$ indexes a source position. A causal window of width $W$ allows entries on the main diagonal and the first $W-1$ subdiagonals. The main diagonal corresponds to the current token. For $T=6$ and $W=3$, the allowed entries are marked by stars.
\[
\begin{bmatrix}
\ast&0&0&0&0&0\\
\ast&\ast&0&0&0&0\\
\ast&\ast&\ast&0&0&0\\
0&\ast&\ast&\ast&0&0\\
0&0&\ast&\ast&\ast&0\\
0&0&0&\ast&\ast&\ast
\end{bmatrix}.
\]
The band determines which source positions can contribute to each output. The mixer determines which entries share parameters, how the entries depend on the input, and whether they are normalized. We use $W=3$ to illustrate the matrices and $W=4$ in the main experiments.

\subsection{From shared diagonals to input-dependent rows}

\paragraph{Depthwise short convolution.} A short convolution copies the same learned coefficient $w_{c,\delta}$ along the diagonal for relative offset $\delta=t-j$. In the example above, the resulting matrix for channel $c$ is
\[
C^{(c)}=
\begin{bmatrix}
w_{c,0}&0&0&0&0&0\\
w_{c,1}&w_{c,0}&0&0&0&0\\
w_{c,2}&w_{c,1}&w_{c,0}&0&0&0\\
0&w_{c,2}&w_{c,1}&w_{c,0}&0&0\\
0&0&w_{c,2}&w_{c,1}&w_{c,0}&0\\
0&0&0&w_{c,2}&w_{c,1}&w_{c,0}
\end{bmatrix}.
\]
This is a causal banded Toeplitz matrix per channel. After training, its signed coefficients are fixed across sequences. The first $W-1$ rows have fewer valid entries, matching left zero padding.

\paragraph{Current-token dynamic convolution.} A dynamic kernel generator can vary the entry on the same diagonal from one row to the next. In our rank-26 implementation~\cite{sieberling2026dynamicshortconvolutionsimprove}, each branch fills row $t$ using the current layer input $h_t$. For every valid offset,
\[
M^{(c)}_{t,t-\delta}=I_{\delta}+b_{c,\delta}+u_{c,\delta}^{\top}D h_t.
\]
Here $D$ projects $h_t$ into the rank-26 space, $u_{c,\delta}$ is a learned vector in that space, and $b_{c,\delta}$ is the static kernel component. The indicator $I_\delta$ is one at offset zero and adds the current payload as a residual connection. The generator emits signed, unnormalized coefficients. Candidate payloads contribute through the weighted sum after the coefficients have been computed.

\paragraph{Sliding-window attention.} Attention fills each allowed entry by scoring its corresponding source position. A softmax over the valid entries of each row gives the coefficient matrix
\[
A(H)=
\begin{bmatrix}
a_{1,1}&0&0&0&0&0\\
a_{2,1}&a_{2,2}&0&0&0&0\\
a_{3,1}&a_{3,2}&a_{3,3}&0&0&0\\
0&a_{4,2}&a_{4,3}&a_{4,4}&0&0\\
0&0&a_{5,3}&a_{5,4}&a_{5,5}&0\\
0&0&0&a_{6,4}&a_{6,5}&a_{6,6}
\end{bmatrix},
\]
where $H$ is the sequence of layer inputs. The band permits the same connections as the convolution. Its entries generally vary along each diagonal, and every row sums to one. Standard sliding-window attention computes each score using the projected query at $t$ and the projected key at $j$.

\subsection{What supplies the query and the keys?}

To connect the formulas, fix one branch, head, and channel or group. Let $K_t$ stack the row vectors used for matching, one per valid offset in row $t$ of the banded matrix. Before any attention scaling or softmax, the row of matching scores is $r_t=q_t^\top K_t^\top$, using a column vector $q_t$. Short convolution uses $q_t=1$ and puts the learned offset scalars in $K_t$. Dynamic convolution uses $q_t=D h_t$ and puts the learned offset vectors in $K_t$. The product yields the dynamic term before adding the static kernel and residual. Standard QK attention derives both operands from the input. SWGDN uses a learned query and input-dependent candidate keys.

In SWGDN, let $\bar p^{X,g}$ and $\bar k_{t-\delta}$ be the learned pseudo-query and the candidate's pre-mixing key RMS normalization. For branch $X$ and group $g$, the logit for offset $\delta$ is
\[
r^{X,g}_{t,\delta}
=\frac{(R_{\delta}\bar p^{X,g})^\top \bar k_{t-\delta}}{\sqrt d}
=\frac{(\bar p^{X,g})^\top R_\delta^\top\bar k_{t-\delta}}{\sqrt d}.
\]
Here $R_\delta$ is the relative RoPE rotation with base 10,000. The second expression places the rotation on the key side, so the matching row of $K_t$ is $(R_\delta^\top\bar k_{t-\delta})^\top$. Each group uses the full key of its head to score candidates, even when it mixes only one payload channel. All three branches share the pre-mixing keys. Each branch applies its own group-specific coefficient matrices to its projected payload, then applies SiLU. The recurrent readout uses the resulting mixed projected Q.

In ordinary QK attention, a change in the current query can change the ordering of two historical candidates with fixed keys. In SWGDN, fixing the two historical keys at a given step fixes their weight ratio $\exp(r^{X,g}_{t,\delta_1}-r^{X,g}_{t,\delta_2})$. A change in the current token's key can still change the softmax denominator and hence the absolute weight of each historical candidate.

\begin{table}[htbp]
\centering
\caption{Query and key operands and coefficient normalization for the four local mixers.}
\label{tab:appendix_mixer_comparison}
\small
\begin{tabular}{@{}llll@{}}
\toprule
Mixer & Query operand & Key operand for offset $\delta$ & Coefficients \\
\midrule
Short convolution & Constant scalar & Learned offset scalar & Signed, direct \\
DynamicConv & Projected current input & Learned offset vector & Signed, direct \\
SWGDN (SWA) & Learned branch/group vector & Candidate key with relative RoPE & Window softmax \\
Input-dependent QK SWA & Projected current query & Projected candidate key & Window softmax \\
\bottomrule
\end{tabular}
\end{table}

\section{Additional Local Mixing Diagnostics}
\label{app:additional_probes}

To characterize how different local mixer designs combine token representations within a window,
we run a series of diagnostic experiments on 340M models with short convolution, DynamicConv, or sliding-window attention.
Section~\ref{app:local_weight_distribution} compares how DynamicConv and sliding-window attention distribute weight across the four offsets in each of the Q, K, and V branches and reports entropy statistics for all three mixers.
Section~\ref{app:key_geometry} examines how well the mixed keys distinguish lexical identity across prefixes and estimates the effective dimensionality of their covariance.

\subsection{Local weight distributions and kernel entropy}
\label{app:local_weight_distribution}

\paragraph{Protocol.}
We compare how sliding-window attention and DynamicConv distribute
weight across the current token and its three predecessors. For each Q, K, and V branch,
we use the sliding-window softmax weights directly. For DynamicConv,
we normalize the absolute effective coefficients as
$p_\delta = |c_\delta|/\sum_j |c_j|$, where
$c=(1+w_0,w_1,w_2,w_3)$ includes the identity term at the current
position. We exclude rows with $\sum_j |c_j| \leq 10^{-12}$.
We pool observations over positions and heads for sliding-window
attention and over positions and channels for DynamicConv, weighting
all 18 layers equally. \Cref{tab:appendix_offset_weights} reports the
mean normalized weight at each offset and how often that offset has
the largest weight.

In both sliding-window models, the current token and its immediate
predecessor together receive 87--94\% of the mean weight in each of
the Q, K, and V branches. The predecessor often receives the largest
weight in V, doing so in 35.0\% of SWKDA-G1 windows and 41.8\% of
SWGDN2-G1 windows.

In DynamicConv's V branch, the current position has the largest
absolute effective coefficient in 96.2\% of position--channel pairs
and receives 77.7\% of the mean normalized coefficient magnitude. In the V branch, DynamicConv concentrates its coefficient magnitude at the current token, and sliding-window attention assigns a larger share of its normalized weight to the preceding token than DynamicConv does.
At the three earlier offsets, the signed coefficients average close
to zero, with 43--55\% negative depending on the branch and offset.

\begin{table}[htbp]
\centering
\caption{Q/K/V weight distributions at all four lags, in percent. DynamicConv statistics use the absolute magnitudes of the effective coefficients.}
\label{tab:appendix_offset_weights}
\small
\setlength{\tabcolsep}{4pt}
\begin{tabular}{@{}llrrrrrrrr@{}}
\toprule
 & & \multicolumn{4}{c}{Argmax frequency (\%)} & \multicolumn{4}{c}{Mean weight (\%)} \\
\cmidrule(lr){3-6}\cmidrule(lr){7-10}
Model & Branch & $t$ & $t-1$ & $t-2$ & $t-3$ & $t$ & $t-1$ & $t-2$ & $t-3$ \\
\midrule
SWKDA-G1 & Q & 93.59 & 5.74 & 0.54 & 0.13 & 77.84 & 15.88 & 3.77 & 2.51 \\
 & K & 62.15 & 31.81 & 4.33 & 1.72 & 55.88 & 32.39 & 8.50 & 3.24 \\
 & V & 60.86 & 35.01 & 3.73 & 0.40 & 55.36 & 35.28 & 7.58 & 1.78 \\
\midrule
SWGDN2-G1 & Q & 69.97 & 29.04 & 0.95 & 0.04 & 61.48 & 31.78 & 5.26 & 1.48 \\
 & K & 66.88 & 27.33 & 4.50 & 1.30 & 57.76 & 30.33 & 8.89 & 3.02 \\
 & V & 52.25 & 41.79 & 5.00 & 0.96 & 47.74 & 39.23 & 10.34 & 2.68 \\
\midrule
GDN2 + DynamicConv & Q & 76.08 & 21.23 & 2.38 & 0.31 & 59.43 & 24.98 & 9.89 & 5.70 \\
 & K & 82.58 & 14.73 & 2.17 & 0.52 & 62.68 & 21.90 & 9.37 & 6.04 \\
 & V & 96.22 & 3.55 & 0.22 & 0.01 & 77.67 & 13.59 & 5.22 & 3.52 \\
\bottomrule
\end{tabular}
\end{table}

We report the mean entropy $\mathbb E[H(p)]$, the entropy of the mean weights $H(\mathbb E[p])$, their Jensen gap, and the mean maximum weight $\mathbb E[\max_\delta p_\delta]$, with $H(p)=-\sum_\delta p_\delta\log p_\delta$. The expectation is over positions within each head and layer for G1, and over position--channel pairs within each layer for DynamicConv. We compute the Jensen gap at that level before averaging. For the KDA and GDN2 short convolutions, we report entropy over the four absolute-normalized learned taps without an added identity, averaging over channels and layers. The uniform reference is $\log 4\approx 1.3863$ nats.

\begin{table}[htbp]
\centering
\caption{Local weight and kernel statistics by model and branch. Entropies and Jensen gaps are in nats. Dashes indicate that only the static-tap entropy is reported. The definitions of normalized weights differ across mixer types.}
\label{tab:appendix_entropy}
\small
\setlength{\tabcolsep}{5pt}
\begin{tabular}{@{}llrrrr@{}}
\toprule
Model & Branch & $\mathbb E[H(p)]$ & $H(\mathbb E[p])$ & Jensen gap & Mean max. \\
\midrule
SWKDA-G1 & Q & 0.5679 & 0.6405 & 0.0726 & 0.8002 \\
 & K & 0.6582 & 0.7677 & 0.1094 & 0.7418 \\
 & V & 0.6083 & 0.7618 & 0.1535 & 0.7639 \\
\midrule
SWGDN2-G1 & Q & 0.6156 & 0.7091 & 0.0935 & 0.7669 \\
 & K & 0.7124 & 0.8195 & 0.1071 & 0.7151 \\
 & V & 0.7295 & 0.8773 & 0.1479 & 0.7054 \\
\midrule
GDN2 + DynamicConv & Q & 0.8437 & 1.0144 & 0.1708 & 0.6710 \\
 & K & 0.8245 & 1.0146 & 0.1901 & 0.6858 \\
 & V & 0.6374 & 0.7364 & 0.0990 & 0.7891 \\
\midrule
KDA & Q & 0.7162 & --- & --- & --- \\
 & K & 0.8332 & --- & --- & --- \\
 & V & 0.7173 & --- & --- & --- \\
\midrule
GDN2 & Q & 0.8754 & --- & --- & --- \\
 & K & 0.9875 & --- & --- & --- \\
 & V & 0.9073 & --- & --- & --- \\
\bottomrule
\end{tabular}
\end{table}

The three local mixers exhibit distinct branch-level patterns
(\cref{tab:appendix_entropy}). In both short-convolution baselines, K has
the highest tap entropy, indicating a more even distribution of absolute
coefficients across the four offsets. In both sliding-window models, V
has the largest Jensen gap and Q the smallest, indicating that the
offset weights vary most across positions in V. In DynamicConv, V has
both the lowest entropy and the smallest Jensen gap, so its normalized
coefficient magnitudes are more concentrated and less variable than
those of Q and K.

Among the G1 models, V has the largest entropy difference (0.6083 nats for SWKDA-G1 and 0.7295 nats for SWGDN2-G1). We hypothesize that channel-wise write gating in GDN2 partly accounts for this difference. When information from neighboring tokens is represented in different value channels, the local mixer can aggregate several sources and the write gate can then regulate which components enter the state. In SWKDA-G1, a single scalar per head scales the entire candidate value. The local mixer may therefore concentrate weight on fewer source positions before the write.

Concentrated weights can still yield similar learning pairs across
consecutive positions. A token can receive a large weight at offset
zero in one window and again at offset one after the window advances.
When this source contributes substantially to both K and V, the two
updates can use related keys and values even if each window assigns
most of its weight to a small number of positions. This distinction
helps explain how SWKDA-G1 can have lower local weight entropy than
SWGDN2-G1 while still producing more similar adjacent learning
pairs than its own short-convolution baseline. In both backbones,
the largest reduction in joint K/V distance occurs at the one-token
offset, where the windows share three of their four source tokens
(\cref{tab:kv_similarity}).

\subsection{Key geometry}
\label{app:key_geometry}
We evaluate mixed keys on 256 synthetic local lexical cases, restricting the identity-gap calculation to the 154 cases where the changed token lies within the four-token window. The identity gap is the mean cosine similarity for the same lexical item under different prefixes minus that for different items under different prefixes, averaged over heads, cases and layers. We also measure the participation ratio $(\sum_i\lambda_i)^2/\sum_i\lambda_i^2$ of centered, L2-normalized key covariance on 64 natural validation blocks, averaging over heads and layers. These are descriptive measurements of separately trained representations.

\begin{table}[htbp]
\centering
\caption{Mixed-key geometry at the final checkpoints. Identity gap and participation ratio are averaged over 18 layers.}
\label{tab:appendix_key_geometry}
\small
\begin{tabular}{@{}lcc@{}}
\toprule
Model & Identity gap & Participation ratio \\
\midrule
GDN2 & 0.0567 & 16.10 \\
GDN2 + DynamicConv & 0.0205 & 14.02 \\
SWGDN2-G1 & 0.0187 & 12.04 \\
SWGDN2-G16 & 0.0186 & 13.41 \\
KDA & 0.0400 & 11.57 \\
SWKDA-G1 & 0.0158 & 12.26 \\
\bottomrule
\end{tabular}
\end{table}

The identity gap is smaller for G1 than its baseline in both backbones (\cref{tab:appendix_key_geometry}), whereas the participation ratio changes in opposite directions.


\end{document}